\pdfoutput=1

\documentclass[11pt]{article}
\usepackage{fourier}
\usepackage{amssymb}
\usepackage{amsfonts}
\usepackage{todonotes}
\usepackage{array}

\usepackage[]{acl}

\usepackage{times}
\usepackage{latexsym}

\usepackage[T1]{fontenc}

\usepackage[utf8]{inputenc}

\usepackage{microtype}

\usepackage{multirow}
\usepackage{varwidth}
\usepackage{cleveref}
\usepackage{booktabs}
\usepackage{xcolor}
\usepackage{tikz}
\usepackage[boldmath]{numprint}
\usepackage{colortbl}
\usetikzlibrary{calc,decorations,shapes,fit,arrows.meta,positioning}

\definecolor{thebluebase}{rgb}{.2,.2,.7}
\colorlet{theblue}{thebluebase!85!white}
\definecolor{thegreen}{rgb}{.53, .66, .42}
\definecolor{theyellow}{rgb}{0.89, 0.66, 0.0}

\usepackage{microtype}

\usepackage{subfig}
\usepackage{multicol}
\usepackage{fancyvrb}
\usepackage{comment}
\usepackage{relsize}

\newsavebox{\verbbox}

\title{Semeval-2022 Task 1: CODWOE -- Comparing Dictionaries and Word Embeddings}

\author{Timothee Mickus\thanks{~~Work conducted while at ATILF} \\
  Helsinki University \\ 
  {} \\
  {\tt timothee.mickus@helsinki.fi} \\\And
  Kees van Deemter \\
  Utrecht University \\
  {\tt c.j.vandeemter@uu.nl} \\\And
  Mathieu Constant \\
  Universit\'e de Lorraine \\
  CNRS, ATILF \\ 
  {\tt mconstant@atilf.fr} \\\And
  Denis Paperno \\
  Utrecht University \\
  {\tt d.paperno@uu.nl} \\}

\begin{document}
\maketitle

\begin{abstract}
    Word embeddings have advanced the state of the art in NLP across numerous tasks.
    Understanding the contents of dense neural representations is of utmost interest to the computational semantics community.
    We propose to focus on relating these opaque word vectors with human-readable definitions, as  found in dictionaries.
    This problem naturally divides into two subtasks: converting definitions into embeddings, and converting embeddings into definitions.
    This task was conducted in a multilingual setting, using comparable sets of embeddings trained homogeneously.
\end{abstract}

\section{Introduction}  \label{sec:intro}

Word embeddings are a success story in \textsc{nlp}. 
They have been equated to distributional semantics models \cite{annurev-lenci-2018,annurev-boleda-2020}, a theory of semantics which relates the meaning of words to their distribution in context  \citep{Harris54}.
Recently introduced contextualized word embeddings \citep[e.g.][]{devlin-etal-2019-bert} have set a new state of the art on a wide variety of tasks. 
For this reason, they have attracted much research interest.
Do they depict consistent semantic spaces and are they theoretically valid \citep{mickus-etal-2020-mean,yenicelik-etal-2020-bert}?
What limitations are to be expected in these models \citep{bender-koller-2020-climbing}? 
Can they scale up in performance \citep{brown2020language}?

Word embeddings are dense vector representations of meaning which are not easily intelligible to a human observer.
Many techniques have been employed to make embedding spaces more interpretable. 
A promising approach consists in \emph{converting these opaque vectors into human readable definitions, as one could find in a dictionary}: accurately translating a dense, opaque vector representation into an equivalent human-readable piece of text would allow us to peer into the black box of modern neural network architectures.
This  avenue of research, known as definition modeling, was pioneered by \citet{DBLP:conf/aaai/NorasetLBD17}.
One may however question whether the task is at all feasible: there is no guarantee that the information content of a dictionary definition is similar to that which is described by real-valued vectors inferred from word distributions. 

\begin{figure}
    \centering
    \includegraphics[width=0.9\linewidth]{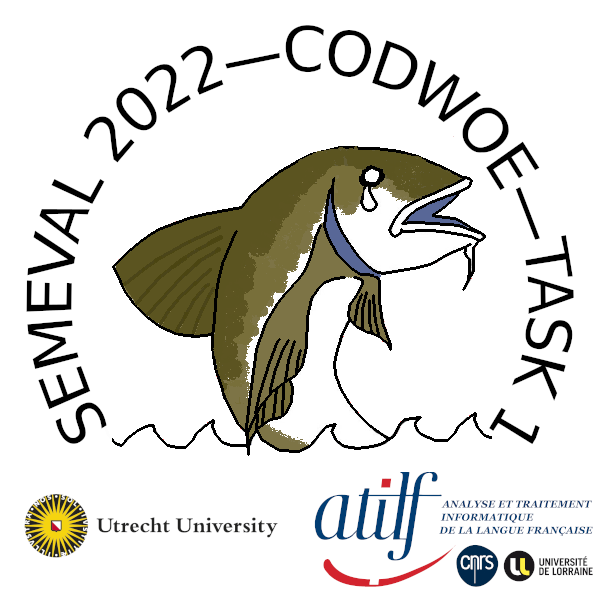}
    \caption{Logo for \textsc{codwoe} shared task}
    \label{fig:logo}
\end{figure}

The SemEval Shared Task on Comparing Dictionaries and Word Embeddings (CODWOE) sets out to study whether embeddings and dictionaries encode similar information.
We present the task and relevant state of the art in \Cref{sec:task}.
We describe the data collected and presented to participants in \Cref{sec:data}.
In \Cref{sec:metrics}, we discuss the metrics used to rank participant submissions.
Our baseline model is presented in \Cref{sec:baselines}.
We list results from participants' submissions in \Cref{sec:results} and provide a more in-depth discussion in \Cref{sec:discussion}.

\section{What we are fishing for} \label{sec:task}

What is in a word embedding? 
Are word embeddings semantic descriptions, in the same sense that dictionary definitions are?
If so, embeddings and definitions must be translatable into one another. The \textsc{codwoe} shared task was set up to test this.
The shared task participants investigated whether a word vector---e.g. $\vec{\texttt{cod}}$---contains the same information as the corresponding dictionary definition---viz. ``\textit{any of various bottom-dwelling fishes (family Gadidae, the cod family) that usually occur in cold marine waters and often have barbels and three dorsal fins}.''\footnote{
    From \href{https://www.merriam-webster.com/dictionary/cod}{Merriam-Webster}.
}

We decompose this research problem into two tracks: the first corresponds to the vector-to-sequence task of Definition Modeling, the second to the sequence-to-vector Reverse Dictionary task.
The task of definition modeling consists in using the vector representation of $\vec{\texttt{cod}}$ to produce the associated gloss, ``\textit{any of various bottom-dwelling fishes (family Gadidae, the cod family) that usually occur in cold marine waters and often have barbels and three dorsal fins}". The reverse dictionary task is the mathematical inverse: reconstruct an embedding $\vec{\texttt{cod}}$ from the corresponding gloss.

These two tracks display a number of interesting characteristics. These tasks are obviously useful for explainable AI, since they involve converting human-readable data into machine-readable data and back. They also have a theoretical significance: both glosses and word embeddings are representations of meaning, and therefore involve the conversion of distinct non-formal semantic representations. From a practical point of view, the ability to infer word-embeddings from dictionary resources, or dictionaries from large un-annotated corpora, would prove a boon for many under-resourced languages.

\subsection{Track 1: Definition Modeling} \label{sec:task:defmod}
The first track consists in an application of Definition Modeling.
As training material, participants have access to a set of data points, each of which consists of a source word embedding and a corresponding target word definition (see \Cref{fig:example-def}).
Participants are tasked with generating new definitions for an unseen test set of embeddings.

Definition Modeling is a recent addition in \textsc{nlg} tasks \citep{DBLP:conf/aaai/NorasetLBD17} which seeks to do just that.
It has since then gained traction 
\cite[a.o.]{gadetsky-etal-2018-conditional,mickus-etal-2019-mark,li-etal-2020-learning,ZHANG2020113633}.
Other languages than English have also been studied, including Chinese \cite{Yang2019Sememes4CDM}, French \cite{mickus-etal-2020-generation}, Wolastoqey \citep{bear-cook-2021-cross}, and more \cite{kabiri2020evaluating}.
At its very inception, Definition Modeling was suggested as a means of evaluating the content of distributional semantic models \citep{DBLP:conf/aaai/NorasetLBD17}.
In practice however, different researchers rarely use comparable sets of embeddings \citep{mickus-etal-2020-generation}, effectively making proper comparisons across systems impossible as they use distinct inputs.
To fill this gap, we created a dataset of comparable embeddings from different languages and neural architectures, trained as homogeneously as possible on comparable data; see \ref{sec-emb-data} below.

\subsection{Track 2: Reverse Dictionary} \label{sec:task:revdict}

Reverse dictionaries (a.k.a. retrograde dictionaries) are lexical resources that flip the usual structure of dictionaries, allowing users to query words based on the definitions they would expect them to have.
One of the major challenges of such resources consists in providing definition glosses that match with users' expectations.
As a consequence, a trend of research in NLP has focused on producing dynamic reverse dictionaries, that would interpret input definitions and map them back to the corresponding word.
We refer the reader to the comprehensive review of \citet{siddique-sufyan-beg-2019-review}, and provide here mainly highlights.

An early strand of research focused on augmenting definitions using synonyms or other semantically related words, such as hypernyms or hyponyms.
This approach has been applied to multiple languages, from Turkish to English and to Japanese \citep{shaw-etal-2013-building,bilac-etal-2004-dictionary,el-khalout-oflazer-2004-use}.
Building on this query-augmentation approach, we find works focused on integrating richer lexical resources, such as WordNet, the Oxford dictionary, The Integral Dictionary, or LDA vector spaces \citep{dutoit-nugues-2002-lexical,thorat-choudhari-2016-implementing,mendez-etal-2013-reverse,calvo-etal-2016-integrated}.

A related trend of research is that of \citet{zanzotto-etal-2010-estimating} and \citet{hill-etal-2016-learning}, who use dictionaries as benchmarks for compositional semantics.
\citet{zanzotto-etal-2010-estimating} used a shallow neural network to implement a compositional distributional semantics model and use dictionaries as their training data.
\citet{hill-etal-2016-learning} instead employ a LSTM to parse the full definition gloss and use the hidden state at the last time-step to predict the word being defined.
In both cases, replacing the definition gloss with a user's query would lead to a reverse dictionary system.
Since then, a number of works have attempted to implement reverse dictionaries using neural language models.
The WantWords system \citep{zhang-2020-multi,qi-2020-wantwords} is based on a BiLSTM architecture, and incorporates auxiliary tasks such as part-of-speech prediction to boost the performance.
\citet{yan-etal-2020-bert} seeks to replace the learned neural language models in \citet{hill-etal-2016-learning} or WantWords with a pre-trained model such as BERT \citep{devlin-etal-2019-bert} and its multilingual variants, which allows them to use their system in a cross-lingual setting---querying in a language to obtain an answer in another.
Most recently, \citet{malekzadeh-etal-2021-predict} used a neural language model based approach to implement a Persian reverse dictionary.

With respect to the CODWOE shared task, our interest lies in reconstructing the word embedding of the word being defined, rather than finding the corresponding word---an approach more closely related to that of \citet{zanzotto-etal-2010-estimating} and \citet{hill-etal-2016-learning}.
Under this slight reformulation, the sequence-to-vector Reverse Dictionary task is strictly the  inverse of the vector-to-sequence task of Definition Modeling. 
Hence we define the Reverse Dictionary task as \emph{computing the components of a target word vector using as input a human-readable definition}.
To solve this task, participants have access to a set of data points, each of which consists of a source word definition and a corresponding target word embedding, as training materials.

\section{What's in the nets: Data used} \label{sec:data}

The definition modeling and reverse dictionary tasks both require a parallel dataset, where dictionary definitions are aligned with corresponding word embeddings. 
The task is held in a multilingual setting. We provide data in English, French, Russian, Italian and Spanish. 
We selected these languages to facilitate the collection of comparable data: all these languages possess comparable large scale resources, including online dictionaries as well as corpora that can be used to train comparable embeddings.
Our datasets are made available online at \url{https://codwoe.atilf.fr/}.

The aim of both tracks of CODWOE is  to compare the semantic contents of definitions and embeddings. As a consequence, we ask participants to refrain from using external data such as pretrained models and lexical resources: including such external data would introduce another source of semantic information, and obfuscate the results from this shared task. 

\subsection{Dictionary data}
As a source of dictionary definitions, we primarily use the \textsc{db}nary dataset \citep{serasset-2012-dbnary},\footnote{
    \url{http://kaiko.getalp.org/about-dbnary/}
} an \textsc{rdf}-formatted version of some of the existing Wiktionary projects.\footnote{
    See \url{https://www.wiktionary.org/}
}
\textsc{Db}nary  includes data for all of our selected languages.
One sub-dataset per language is constructed.
Definitions are selected according to corpus frequency and part-of-speech of the word being defined.
We solely select nouns, adjectives, verbs and adverbs.

\begin{table}[t]
    \centering
    \begin{tabular}{ l  r r }
        \toprule
         & with examples & without \\
        \midrule
        \texttt{en} &      0 & 806297 \\
        \texttt{es} &      0 & 132583 \\ 
        \texttt{fr} & 431793 & 573313 \\
        \texttt{it} &  16127 &  86959 \\
        \texttt{ru} & 122282 & 485208 \\
        \bottomrule
    \end{tabular}
    \caption{\textsc{Db}nary: number of items per language}
    \label{tab:dbnary-items}
\end{table}

Table~\ref{tab:dbnary-items} presents the number of usable items in \textsc{db}nary.
Not all languages contain examples of usage.
A brief regular expression lookup suggests that around 20K examples of usage can be found in the Spanish version of Wiktionary, while English yields at least 200K.
We therefore discard the English version of \textsc{db}nary and replace it by a manual parse, from which we also retrieve examples of usage.


\subsection{Embeddings data}\label{sec-emb-data}
\begin{table}[t]
    \centering
    \begin{tabular}{ l r r r}
        \toprule
          & \textbf{N. Sents.} & \textbf{N. Tokens} & \textbf{N. Bytes} \\ \midrule
         \texttt{it} & 78761031 &  955474050 &  5001829910 \\
         \texttt{es} & 78973969 &  975762257 &  5001999992 \\
         \texttt{fr} & 82082118 & 1004767254 &  5001999368 \\
         \texttt{en} & 97622760 & 1035154295 &  5001999755 \\
         \texttt{ru} & 79526583 & 1035661601 & 10036395727 \\
         \bottomrule
    \end{tabular}
    \caption{Embeddings: corpus statistics}
    \label{tab:cstat}
\end{table}
We have collected similar amounts of data for each language (Table~\ref{tab:cstat}) to use as training corpora.
The sources we use to constitute these corpora are selected to be generally comparable: each corpus contains 2.5G data parsed and cleaned from Wikipedia,\footnote{
    See here: \url{https://dumps.wikimedia.org/}
} 2.2G from the OpenSubtitles OPUS corpus \citep{lison-tiedemann-2016-opensubtitles2016},\footnote{
    See \url{https://opus.nlpl.eu/}
} as well as 0.3G in books from various genres, drawn from LiberLiber\footnote{
    Cf. \url{https://www.liberliber.it/online/}
} for Italian, Wikisource for Spanish and Russian, and Gutenberg\footnote{
    See here: \url{https://www.gutenberg.org/}
} for English and French.

We focus on three embedding architectures: word2vec models \citep{mikolov-etal-2013-efficient}  trained with gensim \citep{rehurek-sojka-2010-software}, the ELECTRA model of \citet{clark-etal-2020-pre}, and character-based embeddings.
The word2vec and ELECTRA models were selected so as to provide some comparison between static and contextual embeddings; both are trained with default hyperparameters aside from output vector size, which we set to 256.
As for the ELECTRA models, given that we need contexts to derive token representations, we train the models only in English, French and Russian. The Spanish and Italian Wiktionary projects contain too few examples of usage.
For French and Russian, we derive contextualized embeddings of a word to be defined from usage examples  in DBnary datasets.
Since the English DBnary dataset does not contain examples of usage, we extracted them from the original Wiktionary dumps.

The character-based embeddings are included to provide baseline expectations for non-semantic representations---as we can expect spelling to be more or less arbitrary with respect to word meaning \citep{saussure-1916-cours}.\footnote{
    Nonetheless, see \citet{gutierrez-etal-2016-finding}, \citet{Kutuzov2017ArbitrarinessOL}, \citet{dautriche17wfsim} or \citet{pimentel-etal-2019-meaning}, all of which question this assumption.
}
In practice, these embeddings are computed through a simple LSTM-based auto-encoder: the word is passed into an LSTM encoder as a sequence of characters, we sum all output hidden states, and use these summed hidden states to initialize an LSTM decoder, whose objective is to reconstruct the input word.
As a character-based representation, we can therefore use the summed output hidden states, as they are tailored to contain all the information necessary to reconstruct the spelling of the corresponding word.\footnote{
    Given that we implement this module ourselves, we use a Bayesian Optimization algorithm \citep{snoek-etal-2012-practical} to select hyperparameters for our five character auto-encoder.
    We use this process to decide learning rate, weight decay, dropout, $\beta_1$ and $\beta_2$ parameters of the AdamW optimizer, batch size, number of epochs over the full dataset, as well as whether to share a single weight matrix for encoder and decoder character embeddings.
}
The datasets used to trained the models correspond to the set of all word types attested in our base corpora described in \Cref{tab:cstat}.
All models achieve a 99\% reconstruction accuracy.

\subsection{Datasets}
We construct one dataset per language.
Each language-specific dataset is split in five: a trial split (200 datapoints per language), a training split (43~608 datapoints), a validation split (6375 datapoints), a definition modeling testing split (6221 datapoints) and a reverse-dictionary testing split (6208 datapoints). 
Splits are constructed such that there are no overlap in the embeddings.
Dataset splits are formatted as \textsc{json} files.

Each file consists of a list of \textsc{json} dictionary notations.
\textsc{Json} items contain a unique identifier for the data point, the word being defined, definition, part of speech, and all word vectors.
A depiction of the sort of items included in our datasets is shown in \Cref{fig:example-def}.
Sub-figure~\ref{fig:example-def:human-readable} summarizes the data presented as a \textsc{json} item in Sub-figure~\ref{fig:example-def:json}.

\begin{lrbox}{\verbbox}
\begin{minipage}{.5\textwidth}
\begin{verbatim}
{
    "id": "it.42",
    "word": "sminuire"
    "gloss": "far figurare...",
    "pos": "v",
    "electra": [0.4, 0.2, ...],
    "sgns": [0.2, 0.4, ...],
    "char": [0.3, 1.4, ...],
}
\end{verbatim}
\end{minipage}
\end{lrbox}

\begin{figure}[t]
    \subfloat[\label{fig:example-def:human-readable} Example definition in Italian]{
        \begin{tabular}{l@{{~}}c@{{~}}p{5cm}}
            \toprule
            \textbf{word} & \textbf{\textsc{pos}} & \textbf{gloss}  \\ 
            \midrule
            sminuire & V & far figurare qualcosa o qualcuno come meno importante o rilevante \\ 
            \bottomrule
        \end{tabular}
    }
    
    \subfloat[\label{fig:example-def:json} Corresponding \textsc{json} snippet]{
        \usebox{\verbbox}
    }
    \caption{Toy example data point in the Italian dataset}
    \label{fig:example-def} 
\end{figure}

Participants had access to the trial, train and validation splits of all languages.
Test splits were made available at the beginning of the evaluation period.

\section{The scales we use} \label{sec:metrics}

We now turn to the metrics of our shared task.
\subsection{Reverse Dictionary Metrics}
The Reverse Dictionary task, as we have re-framed it here, consists in reconstructing embeddings.
To that end, we consider three measures of vector similarity.
First is MSE (mean squared error), which measures the difference between the components of the reconstructed and target embeddings. Mean-squared error is however not very easy to interpret on its own.
Second is cosine: the reconstructed and target embeddings should have a cosine of 1.
It is hard to place specific expectations for what a random output would produce, as this essentially differs from architecture to architecture: for instance, Transformer outputs are known to be anisotropic, so we shouldn't expect two random ELECTRA embeddings to be orthogonal \citep[a.o.]{ethayarajh-2019-contextual,timkey-van-schijndel-2021-bark}.

As neither MSE nor cosine provides us with a clear diagnosis tool comparable across all targets, we also include a ranking based measure: we compare the cosine of the reconstructed embedding $\vec{p}_i$ and the target embedding $\vec{t}_i$ to the cosine of the reconstruction $\vec{p}_i$ and all other targets $\vec{t}_j$ in the test set, and evaluate the proportion of such targets that would yield a closer association---viz., the number of cosine values greater than $\cos(\vec{p}_i , \vec{t}_i)$.
More formally, we can describe this ranking metric as: 
\begin{equation} \label{eq:codwoe:ranking}
	\textrm{Ranking}(\vec{p}_i) = \frac{
		\sum\limits_{
			\vec{t}_j \in \textrm{Test set}
		}
		\mathbb{1}_{
			\cos(\vec{p}_i , \vec{t}_j) > \cos(\vec{p}_i , \vec{t}_i)
		}
	}
	{
		\#  \textrm{Test set}
	}
\end{equation}

\subsection{Definition Modeling Metrics}
A common trope in NLG is to stress the dearth of adequate automatic metrics.
Most of the metrics currently existing focus on token overlap, rather than semantic equivalence.
The very popular BLEU and ROUGE metrics \citep{papineni-etal-2002-bleu,lin-2004-rouge} measure the overlap rate in n-grams of various lengths (usually 1-grams to 4-grams).

To alleviate this, researchers have suggested using external resources, such as lists of synonyms and stemmers \citep{banerjee-lavie-2005-meteor} or pre-trained language models \citep{zhao-etal-2019-moverscore}.
The reliance of these augmented metrics on external resources is problematic.
Different languages will use different resources with varying degrees of quality---and this will necessarily impact scores, introducing a confounding factor for any analysis down the line.
In the extreme case, if these resources are not available for a particular language, then the metric will have to be discarded.
Even assuming the availability of the required external resources, none of these improved metrics is entirely satisfactory.
In the case of synonymy-aware metrics such as METEOR \citep{banerjee-lavie-2005-meteor}, we can stress that syntactically different sentences can express the same meaning, but would not be captured by such metrics.
Embeddings-based metrics such as MoverScore \citep{zhao-etal-2019-moverscore} are very recent, and therefore less well understood; moreover concerns can be raised about whether using a method derived from neural networks trained on text will prove of any help in studying the meaning of texts generated by other neural networks.

One alternative frequently used by the NLG community is perplexity, which weighs the probability that the model would generate the target.
This last alternative is however not suited to a shared task setup, as it requires us to have access to the actual neural networks trained by participants so we can investigate the probability distributions they model---unlike the other metrics we mentioned thus far, which only require model outputs.

In short, none of the currently available NLG metrics are fully satisfactory.
Some are not applicable given the shared task format, some depend on external resources of varying quality, and some merely measure formal similarity, rather than semantic equivalence.
Our approach is therefore twofold: on the one hand, we select multiple metrics with the expectation that each might shed light on one specific factor; on the other hand, we encourage participants to go beyond automatic scoring for the evaluation of their model.

As for which metrics we select, we narrow our choice to three.
First is a basic BLEU score \citep{papineni-etal-2002-bleu} between a production $p_i$ and the associated target $t_i$; our reasoning here is that as it is one of the most basic metrics, it is a consistent default choice.
Second is the maximum BLEU score between a production $p_i$ and any of the targets $t_i, t_j \dots t_n$ for which the definiendum is the same as that of $p_i$.
This second metric is designed to not penalize models that rely solely on SGNS or char embeddings: as the input would always be the same, deterministic models would always produce the same definition $p_i = p_j =\dots = p_n$.\footnote{
	One way of bypassing this problem would be to include a source of noise, as is done in GAN architectures \citep{NIPS2014_5ca3e9b1}.
	This would still leave open the question of how to optimally align the outputs to the possible targets.
}
To distinguish between our two BLEU variants, we refer to the former as S-BLEU (or Sense-BLEU), and the latter as L-BLEU (or Lemma-BLEU).

Given that some definitions in our dataset can be very short, we also apply a smoothing to both BLEU-based metrics.
In practice, BLEU computes an overlap of n-grams of size $m$ and under; by default, $m=4$.
This overlap is a geometric mean across all n-gram sizes $1 \dots m$.
If a definition $d$ contains less than $m$ tokens, then any associated production for which $d$ is used as a target will contain 0 overlapping n-grams of size $m$.
The use of a geometric mean then entails that the BLEU score for any production associated to $d$ will be 0.
To circumvent this limitation of BLEU, it is common to use some form of smoothing.
Here, for any n-gram size $\hat{m}$ that would yield an overlap of 0 (i.e., $\hat{m}$ such that $\#d < \hat{m} \leq m$), we replace the overlap count with a pseudocount of $1/\log \#d$.

Lastly, we include MoverScore \citep{zhao-etal-2019-moverscore}, using a multilingual DistilBERT model as the external resource.
The fact that this model is multilingual means that we can use it for all five languages of interest.
Embedding-based methods have the potential to overcome some of the limitations of purely token-based metrics, which is why we deem them worth including in our setup.

The second part of our approach for evaluating submissions consists in encouraging participants to not rely solely on the automatic scoring system of their outputs.
Concretely, we provide participants with a richly annotated trial dataset, which contains frequency and hand-annotated semantic information, and strongly suggest participants to use it for a manual evaluation of their system.
We include the presence of a manual evaluation as a criterion to evaluate the quality of a system description paper, and plan to formally recognize the most enlightening evaluations conducted by participants.

Neither our selection of metrics nor our insistence on manual evaluation solves the evaluation issues of NLG systems.
We duly note the importance of this question, and plan to conduct a follow-up evaluation campaign on the CoDWoE submissions.

\section{Testing the waters: baseline architectures} \label{sec:baselines}
We implement simple neural network architecture baselines to lower the barrier to entry to this shared task. 
They are based on the Transformer architecture of \citet{vaswani-etal-2017-attention} and designed to be as simple as possible.
Our code is publicly available at \url{https://github.com/TimotheeMickus/codwoe}.

\begin{figure}[t]
	\centering
	\subfloat[
	\label{fig:codwoe:revdict-baseline-arch}
	Reverse dictionary
	]{
	\begin{tikzpicture}[
	decstyle/.style={rectangle,draw,fill=white!65!red},
	scale=1]
	\foreach \x in {0,...,4}
	\foreach \y in {0,...,3}
	{\node [decstyle] (d\x\y) at (.6*\x, .6*\y) {};}

	\node (sum) [circle,draw,inner sep=-0.5pt,minimum height =.2cm,] at (.6*2,.6 * 4.5  + 0.2) {$+$};
	\node (pred) at (.6*2,.6 * 6 + 0.2) {$p_i$};
	\draw[-stealth,gray] (sum.north)--(pred.south);

	\foreach \y [count=\yi] in {0,...,2}
	\foreach \x in {0,...,4}
	\foreach \xi in {0,...,4}
	\draw[-stealth,gray] (d\x\y.north)--(d\xi\yi.south);

	\foreach \t [count=\ti from 0] in {{$\vec{\texttt{bos}}$}, {$\vec{w_1}$}, {$\dots$}, {$\vec{w_n}$}, {$\vec{\texttt{eos}}$}}
	{
		\node[rotate=55] (e\ti0) at (.6*\ti,-1.31) {\t};
	}
	\foreach \x in {0,...,4} {
		\draw[-stealth,gray] (e\x0)--(d\x0.south);
		\draw[-stealth,gray] (d\x3.north)--(sum);
	}
	\end{tikzpicture}
}
\subfloat[
	\label{fig:codwoe:defmod-baseline-arch}
	Definition modeling
]{
	\begin{tikzpicture}[
	decstyle/.style={rectangle,draw,fill=white!65!thegreen},
	scale=1]
     \foreach \x in {0,...,4}
     \foreach \y in {0,...,3}
      \node[decstyle] (d\x\y) at (.6*\x + 6,.6 *\y) {};
     \foreach \y [count=\yi] in {0,...,2}
     \foreach \x in {0,...,4}
     \foreach \xi in {\x,...,4}
     \draw[-stealth,gray, ultra thin] (d\x\y.north)--(d\xi\yi.south);

    \foreach \t [count=\ti from 0] in {{\texttt{bos}}, {$p^1_i$}, {$\dots$}, {$p^n_i$}, {\texttt{eos}}}
     {
     \node[rotate=75] (d\ti4) at (.6*\ti + 6,.6 * 4 + 0.6) {\t};
     }
     \foreach \x in {0,...,4} \draw[-stealth,gray] (d\x3.north)--(d\x4);

     \foreach \t [count=\ti from 0] in {{$\vec{d_i}$}, {$\vec{\texttt{bos}}$}, {$\vec{w_1}$}, {$\dots$}, {$\vec{w_n}$}}
     {
        \node[rotate=75] (id\ti0) at (.6*\ti + 6,-1.25) {\t};
        \draw[-stealth,gray] (id\ti0)--(d\ti0.south);
     }
	\end{tikzpicture}
	}
	\caption{Baseline architectures for the CoDWoE shared task}
	\label{fig:codwoe:baseline-archs}
\end{figure}
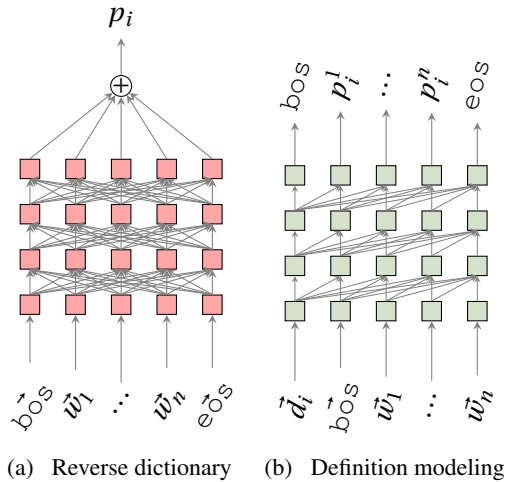

We illustrate our Reverse Dictionary baseline architecture in \Cref{fig:codwoe:revdict-baseline-arch}.
It consists in feeding the input gloss $\langle \vec{\texttt{bos}}, \vec{w_1}, \dots, \vec{w_n}, \vec{\texttt{eos}} \rangle$ into a simple Transformer encoder, and then summing all the hidden representations to produce the prediction $p_i$.
In practice, the summed hidden states are passed into a small non-linear feed-forward module to derive the prediction:
\begin{equation}
		p_i = W_p \left( \textrm{ReLU}\left(\sum\limits_{t} \vec{h_t}\right) \right)
\end{equation}

Our Definition Modeling baseline is presented in \Cref{fig:codwoe:defmod-baseline-arch}.
It consists in a simple Transformer encoder, where earlier time-step representations are prevented from attending to later time-step representations.
To provide information about the definiendum to the model, we use the definiendum embedding $\vec{d_i}$ as the input for the first time-step instead of a start-of-sequence token.
We train the models with teacher-forcing: i.e., during training we ignore the definientia $p^1_i, \dots, p^n_i$ that the model produces; instead we feed it the target $w_1, \dots, w_m$ attested in the training set at each time-step.
During inference, we feed the model with its own prediction.
This creates a train-test mismatch, which we alleviate by using a beam-search.
We stop generation when all beams have produced an end-of-sequence token.

\begin{table*}[t]
    \centering
    \setlength{\aboverulesep}{0pt}
\setlength{\belowrulesep}{0pt}
\setlength{\extrarowheight}{.75ex}
\resizebox{\textwidth}{!}{
    \nprounddigits{3}%
    \npdecimalsign{.}%
    \begin{tabular}{l | n{1}{3} n{0}{3} n{0}{3} >{\columncolor{lightgray}}n{1}{3} >{\columncolor{lightgray}}n{0}{3} >{\columncolor{lightgray}}n{0}{3} n{1}{3} n{0}{3} n{0}{3} >{\columncolor{lightgray}}n{1}{3} >{\columncolor{lightgray}}n{0}{3} >{\columncolor{lightgray}}n{0}{3} n{1}{3} n{0}{3} n{0}{3}}%
        \toprule%
        \multirow{2}{*}{\textbf{Team}} & \multicolumn{3}{c}{\texttt{en}} & \multicolumn{3}{>{\columncolor{lightgray}}c}{\texttt{es}} & \multicolumn{3}{c}{\texttt{fr}} & \multicolumn{3}{>{\columncolor{lightgray}}c}{\texttt{it}} & \multicolumn{3}{c}{\texttt{ru}}  \\%
        & {{Mv}} & {{SB}} & {{LB}} & {{Mv}} & {{SB}} & {{LB}} & {{Mv}} & {{SB}} & {{LB}} & {{Mv}} & {{SB}} & {{LB}} & {{Mv}} & {{SB}} & {{LB}} \\
        \midrule%
Bl. SGNS & 0.084 & 0.030 & 0.040 & 0.065 & 0.035 & 0.052 & 0.046 & 0.030 & 0.041 & 0.107 & 0.053 & 0.076 & 0.112 & 0.039 &  0.054 \\
Bl. char & 0.047 & 0.026 & 0.033 & 0.059 & 0.031 & 0.043 & 0.022 & 0.028 & 0.037 & 0.046 & 0.029 & 0.038 & 0.072 & 0.025 & 0.037 \\
        Bl. Electra      & 0.06498386286509694   & 0.030857978973476703  & 0.03940406882392039  &                       &                       &                       & 0.0431985290276657    & 0.031283591619142584  & 0.039112105271958755 &  & & & 0.10075702410093894 & 0.0318996380927413 & 0.04125104599833407 \\%
        \midrule%
        Locchi           & 0.04898               & 0.02224               & 0.0272               & 0.03776               & 0.02007               & 0.02623               &                       &                       &         & 0.07124 & 0.00788 & 0.01222 &  &  &  \\ %
        LingJing         & -0.04507              & 0.00365               & 0.00505              & 0.02284               & 0.01258               & 0.01956               & -0.11322              & 0.00303               & 0.00453 & -0.01241 & 0.0179 & 0.02915 & -0.01032 & 0.0109 & 0.01426 \\ %
        BLCU-ICALL       & {\npboldmath} 0.13475 & 0.03127               & 0.03957              & {\npboldmath} 0.12778 & 0.03914               & 0.05606               & 0.04193               & 0.02679               &0.03691 & {\npboldmath} 0.11717 & {\npboldmath} 0.06646 &  {\npboldmath}  0.09926 &  {\npboldmath} 0.1482 & 0.04843 & 0.06548 \\ %
        IRB-NLP          & 0.09419               & {\npboldmath} 0.03334 & 0.0417               & 0.09323               & {\npboldmath} 0.04467 & {\npboldmath} 0.06443 & 0.05621               & 0.02787               & 0.03319 & 0.07698 & 0.01045 & 0.01547 & 0.07996 & 0.02721 & 0.03612 \\ %
        RIGA             & 0.0928                & 0.02564               & 0.03153              & 0.10679               & 0.03057               & 0.0445                & {\npboldmath} 0.07468 & 0.02449               & 0.02985 & 0.09337 & 0.0122 & 0.01824 & 0.09387 & 0.03127 & 0.04251 \\ %
        lukechan1231     & 0.07126               & 0.02154               & 0.02658              & 0.06776               & 0.02502               & 0.03599               & 0.05351               & 0.02076               & 0.02571 & 0.1011 & 0.03742 & 0.05353 & 0.109 & 0.02864 & 0.03964 \\ %
        Edinburgh        & 0.10368               & 0.03055               & 0.0381               & 0.10061               & 0.03467               & 0.05275               & 0.0263                & {\npboldmath} 0.02946 &  {\npboldmath} 0.03824 & 0.10677 & 0.0601 & 0.09167 & 0.10925 &  {\npboldmath} 0.04908 &  {\npboldmath} 0.07235 \\ %
        talent404        & 0.12847               & 0.03278               & {\npboldmath} 0.0425 &  &  &  &  &  &  &  &  &  &  &  &  \\ %
        \bottomrule%
    \end{tabular}%
}
    \caption{\label{tab:res-defmod}Participants' best scores on the Definition Modeling track. Highest participant scores per metric are displayed in bold font.}
\end{table*}

For both tracks, we train one model for each distinct pair of language and embedding architecture.
We start by re-tokenizing the datasets using sentence piece with a vocabulary size of 15000.
This is done in order to mitigate the effects of different vocabulary sizes when training our Transformer baselines, and make the models overall easier to compare across different languages.

We set hyperparameters using a Bayesian Optimization procedure, with 100 hyperparameter configurations tested and 10 initial random samples.
For the Reverse dictionary models, we tune the following hyper-parameters: learning rate, weight decay penalty, the $\beta_1$ and $\beta_2$ hyperparameters of the Adam optimizing algorithm, dropout rate, length of warmup, batch size,\footnote{
	In practice, we first manually find the largest batch size that fits on our GPU, and then let the model select the number of batches it should accumulate gradient on.
} number of heads in the multi-head attention layers, and number of stack layers.
For the Definition Modeling systems, we also include a label smoothing parameter to tune.
Models are trained over up to 100 epochs; training is stopped early if no improvement of at least $0.1\%$ is observed during 5 epochs.
In all cases, we decay the learning rate after the warmup following a half cosine wave, such that the learning rate reaches 0 at the end of the 100 epochs.

\begin{table*}[!h]
    \centering
    \subfloat[\label{tab:res-revdict-sgns}SGNS Reverse Dictionary track results]{%
        \setlength{\aboverulesep}{0pt}
\setlength{\belowrulesep}{0pt}
\setlength{\extrarowheight}{.75ex}
\resizebox{\textwidth}{!}{
    \nprounddigits{3}%
    \npdecimalsign{.}%
    \smaller
    \begin{tabular}{l | n{1}{3} n{0}{3} n{0}{3} >{\columncolor{lightgray}}n{1}{3} >{\columncolor{lightgray}}n{0}{3} >{\columncolor{lightgray}}n{0}{3} n{1}{3} n{0}{3} n{0}{3} >{\columncolor{lightgray}}n{1}{3} >{\columncolor{lightgray}}n{0}{3} >{\columncolor{lightgray}}n{0}{3} n{1}{3} n{0}{3} n{0}{3}}%
        \toprule%
        \multirow{2}{*}{\textbf{Team}} & \multicolumn{3}{c}{\texttt{en}} & \multicolumn{3}{>{\columncolor{lightgray}}c}{\texttt{es}} & \multicolumn{3}{c}{\texttt{fr}} & \multicolumn{3}{>{\columncolor{lightgray}}c}{\texttt{it}} & \multicolumn{3}{c}{\texttt{ru}}  \\%
        & {{MSE}} & {{cos}} & {{rnk}} & {{MSE}} & {{cos}} & {{rnk}} & {{MSE}} & {{cos}} & {{rnk}} & {{MSE}} & {{cos}} & {{rnk}} & {{MSE}} & {{cos}} & {{rnk}} \\
        \midrule%
        Baseline & 0.9109240770339966 & 0.15132397413253784 & 0.49029124151800096 & 0.9299595355987549 & 0.20405696332454681 & 0.4991232881840971 & 1.1405054330825806 & 0.19773952662944794 & 0.4905176064402787 & 1.1253628730773926 & 0.20430098474025726 & 0.47691825001510146 & 0.5768295526504517 & 0.2531660497188568 & 0.4900802926918895\\%
        \midrule%
Locchi & 0.87462 & 0.20447 & 0.39422 &  &  &  &  &  &  & 1.08749 & 0.2744 & 0.38598 &  &  &  \\ %
BL.research & 0.89506 & 0.16551 & 0.31194 & 0.91046 & 0.25195 & 0.25342 & 1.10697 & 0.21162 & 0.31376 & 1.11142 & 0.24558 & 0.2468 & 0.56629 & 0.29813 & 0.28952 \\ %
LingJing & 0.8624 & 0.24311 & 0.32907 & {\npboldmath} 0.85771 & 0.35275 & 0.25102 & 1.02969 & 0.32799 & 0.28214 & 1.03946 & 0.35955 & 0.22995 & {\npboldmath} 0.52827 & {\npboldmath} 0.42441 & 0.18712 \\ %
MMG &  &  &  & 0.91068 & {\npboldmath} 0.40293 & {\npboldmath} 0.1665 &  &  &  &  &  &  &  &  &  \\ %
chlrbgus321 & {\npboldmath} 0.85397 & 0.24786 & 0.31857 &  &  &  &  &  &  &  &  &  &  &  & \\ %
IRB-NLP &  0.96367 & {\npboldmath} 0.25968 & {\npboldmath} 0.23126 & 0.88337 & 0.36731 & 0.19673 & 1.06834 & {\npboldmath} 0.34156 & {\npboldmath} 0.19292 & 1.07601 & {\npboldmath} 0.38013 & {\npboldmath} 0.16469 & 0.56834 & 0.4211 & {\npboldmath} 0.15047 \\ %
Edinburgh & 0.86399 & 0.24065 & 0.32642 & 0.85959 & 0.34661 & 0.27096 & {\npboldmath} 1.0259 & 0.31181 & 0.30212 & {\npboldmath} 1.03092 & 0.37395 & 0.19661 & 0.53768 & 0.38278 & 0.24675 \\ %
the0ne & 0.89953 & 0.18531 & 0.50043 &  &  &  &  &  &  &  &  &  &  &  &  \\ %
JSI & 0.90939 & 0.15578 & 0.49891 & 0.91334 & 0.22315 & 0.49469 & 1.12245 & 0.21621 & 0.49772 & 1.19625 & -0.004 & 0.49871 & 0.6151 & 0.00568 & 0.49853 \\ %
1cadamy  & 0.91468 & 0.1939 & 0.37361 & 0.90623 & 0.26175 & 0.37545 & 1.09962 & 0.22757 & 0.43902 & 1.09696 & 0.25959 & 0.38376 & 0.57808 & 0.33516 & 0.29077 \\ %
        \bottomrule%
    \end{tabular}%
}
    }%
    
    \subfloat[\label{tab:res-revdict-char}Char Reverse Dictionary track results]{%
        \setlength{\aboverulesep}{0pt}
\setlength{\belowrulesep}{0pt}
\setlength{\extrarowheight}{.75ex}
\resizebox{\textwidth}{!}{
    \nprounddigits{3}%
    \npdecimalsign{.}%
    \smaller
    \begin{tabular}{l | n{1}{3} n{0}{3} n{0}{3} >{\columncolor{lightgray}}n{1}{3} >{\columncolor{lightgray}}n{0}{3} >{\columncolor{lightgray}}n{0}{3} n{1}{3} n{0}{3} n{0}{3} >{\columncolor{lightgray}}n{1}{3} >{\columncolor{lightgray}}n{0}{3} >{\columncolor{lightgray}}n{0}{3} n{1}{3} n{0}{3} n{0}{3}}%
        \toprule%
        \multirow{2}{*}{\textbf{Team}} & \multicolumn{3}{c}{\texttt{en}} & \multicolumn{3}{>{\columncolor{lightgray}}c}{\texttt{es}} & \multicolumn{3}{c}{\texttt{fr}} & \multicolumn{3}{>{\columncolor{lightgray}}c}{\texttt{it}} & \multicolumn{3}{c}{\texttt{ru}}  \\%
        & {{MSE}} & {{cos}} & {{rnk}} & {{MSE}} & {{cos}} & {{rnk}} & {{MSE}} & {{cos}} & {{rnk}} & {{MSE}} & {{cos}} & {{rnk}} & {{MSE}} & {{cos}} & {{rnk}} \\
        \midrule%
        Baseline & 0.14775659143924713 & 0.7900623679161072 & 0.5021822034698171 & 0.5695170164108276 & 0.8063414096832275 & 0.4977826383924976 & 0.3948018550872803 & 0.7585152983665466 & 0.49945084090085373 & 0.3630896806716919 & 0.7273207902908325 & 0.4966306391450548 & 0.13498325645923615 & 0.8262424468994141 & 0.494515212540774\\%
        \midrule%
        Locchi &  {\npboldmath} 0.14142 & {\npboldmath} 0.79764 & 0.48334 &  &  &  &  &  &  & 0.35457 & 0.73368 & 0.47833 &  &  &  \\ %
        BL.research & 0.14307 & 0.79532 & 0.44967 & 0.51047 & 0.82412 & 0.41168 & 0.36585 & 0.76985 & 0.42815 & 0.3587 & 0.72812 & 0.41702 & 0.13215 & 0.82962 & 0.40971 \\ %
        LingJing & 0.17603 & 0.78204 & 0.48599 & 0.58332 & 0.82366 & 0.50036 & 0.41096 & 0.75165 & 0.50159 & 0.43755 & 0.68085 & 0.49603 & 0.18359 & 0.79058 & 0.47229 \\ %
        IRB-NLP & 0.16156 & 0.77004 & {\npboldmath}0.41874 & 0.52623 & 0.81927 & {\npboldmath} 0.40295 & 0.3896 & 0.75627 & 0.42142 & 0.36643 & 0.72416 & {\npboldmath} 0.38324 & 0.14018 & 0.82441 & {\npboldmath} 0.35675 \\ %
        Edinburgh & 0.14292 & 0.7954 & 0.4997 & {\npboldmath} 0.4667 & {\npboldmath} 0.83882 & 0.42441 & {\npboldmath} 0.33539 & {\npboldmath} 0.78853 & 0.42833 & {\npboldmath}  0.33434 & {\npboldmath} 0.7473 & 0.42843 & {\npboldmath}  0.11613 & {\npboldmath} 0.85174 & 0.38936 \\ %
        the0ne & 0.14337 & 0.79565 & 0.50043 &  &  &  &  &  &  &  &  &  &  &  &  \\ %
        1cadamy  & 0.16764 & 0.79247 & 0.47809 & 0.55655 & 0.8196 & 0.40979 & 0.39075 & 0.76861 & {\npboldmath} 0.41623 & 0.36354 & 0.73924 & 0.43793 & 0.15608 & 0.83637 & 0.37722 \\ %
        \bottomrule%
    \end{tabular}%
}
    }%

    \subfloat[\label{tab:res-revdict-electra}ELECTRA Reverse Dictionary track results]{%
        \setlength{\aboverulesep}{0pt}
\setlength{\belowrulesep}{0pt}
\setlength{\extrarowheight}{.75ex}
\resizebox{0.66\textwidth}{!}{
    \nprounddigits{3}%
    \npdecimalsign{.}%
    \smaller
    \begin{tabular}{l | n{1}{3} n{0}{3} n{0}{3} >{\columncolor{lightgray}}n{1}{3} >{\columncolor{lightgray}}n{0}{3} >{\columncolor{lightgray}}n{0}{3} n{1}{3} n{0}{3} n{0}{3}}%
        \toprule%
        \multirow{2}{*}{\textbf{Team}} & \multicolumn{3}{c}{\texttt{en}} & \multicolumn{3}{>{\columncolor{lightgray}}c}{\texttt{fr}} & \multicolumn{3}{c}{\texttt{ru}} \\%
        & {{MSE}} & {{cos}} & {{rnk}} & {{MSE}} & {{cos}} & {{rnk}} & {{MSE}} & {{cos}} & {{rnk}} \\
        \midrule%
        Baseline & 1.4128680229187012 & 0.8428266048431396 & 0.49848694162270457 & 1.1534641981124878 & 0.8562874794006348 & 0.49783513963836984 & 0.8735798597335815 & 0.7208637595176697 & 0.491197763030062 \\%
        \midrule%
        Locchi & {\npboldmath} 1.30125 & 0.8427 & 0.47784 &  &  &  &  &  & \\ %
        BL.research & 1.32593 & 0.84368 & 0.4339 & 1.1115 & 0.85806 & 0.44182 & 0.86408 & 0.72063 & 0.39866\\ %
        LingJing & 1.50876 & 0.84592 & 0.47773 & 1.27067 & 0.85859 & 0.47763 & {\npboldmath} 0.82774 & 0.73397 & 0.42021\\ %
        IRB-NLP & 1.68464 & 0.82842 & {\npboldmath} 0.4315 & 1.33862 & 0.84664 & {\npboldmath} 0.42921 & 0.91132 & 0.72363 & {\npboldmath} 0.34478\\ %
        Edinburgh & 1.31049 & {\npboldmath} 0.84667 & 0.49042 &{\npboldmath} 1.06591 & {\npboldmath} 0.86171 & 0.4756 & 0.82848 & {\npboldmath} 0.7352 & 0.41711\\ %
        the0ne & 1.34014 & 0.84564 & 0.50008 &  &  &  &  &  & \\ %
        \bottomrule%
    \end{tabular}%
}
    }%
    \caption{\label{tab:res-revdict} Participants' best scores on the Reverse Dictionary track. Highest participant scores per metric are displayed in bold font.}
\end{table*}

\section{How whale did it go? Shared task results.} \label{sec:results}

Scores attained by participants are shown in \Cref{tab:res-defmod,tab:res-revdict}.
In \Cref{tab:res-defmod}, ``Mv'', ``SB'' and ``LB'' refer to Moverscore, Sense-BLEU and Lemma-BLEU respectively; in \Cref{tab:res-revdict}, each sub-table corresponds to a different architecure, and ``rnk'' refers to the cosine ranking metric (cf. \Cref{sec:metrics}). 

In total, we received 159 valid submissions from 15 different users; out of which 11 teams produced a submission paper. 
9 of these teams tackled the Definition Modeling, and 10 addressed the reverse dictionary track.
Competition rankings are established by ranking each submission received, selecting for each participant the best performance on all metrics, and finally taking the average best rank. 
Some participants' submissions were faulty and could not be processed by the evaluation website scoring program.

Among the system descriptions we received, two focused solely on definition modeling.
\citet[BLCU-ICALL]{submissions-blcuicall} use a multitasking framework for definition modeling, based on a generation and a reconstruction objectives. 
\citet[RIGA]{submissions-riga} focus on what are the effects of model size and duration of training on GRUs and LSTMs for definition modeling, and whether MoverScore corroborates human judgment.

Five submissions specifically focus on the reverse dictionary task.
\citet[BL.research]{submissions-blresearch} compare the performances of MLP-based to LSTM-based networks for reverse dictionary. 
\citet[LingJing]{submissions-lingjing} study pretraining objectives for the reverse dictionary track.
\citet[MMG]{submissions-mmg} pay specific attention to how the not-so-satisfactory quality of the Spanish dataset impacts results on Spanish reverse dictionary.
\citet[Uppsala]{submissions-uppsala} study whether foreign language entries can improve the performance of the English reverse dictionary baseline model.
\citet[1cademy]{submissions-zhwa3087} introduce multiple 
technical tweaks for reverse dictionary, such as a dynamic weight averaging loss, language-specific tags and residual cutting.

The last four submissions addressed both tracks.
\citet[Edinburgh]{submissions-edinburgh} propose to project embeddings and definitions on a shared representational space.
\citet[IRB-NLP]{submissions-irbnlp} take inspiration from \citet{DBLP:conf/aaai/NorasetLBD17} to address definition modeling, and experiment with pooling strategies over Transformer embeddings for the reverse dictionary track.
\citet[JSI]{submissions-jsi} focus on comparing the effects of adding LSTM and BiLSTM layers on top of a Transformer model, as well as zero-shot cross-lingual generalization.
\citet[TLDR]{submissions-tldr} propose two Transformer-based architectures for the two tracks, leveraging contrastive learning and unsupervised pretraining.

Looking at \Cref{tab:res-defmod,tab:res-revdict}, we see that the metrics we chose in \cref{sec:metrics} are not always aligned.
On the Definition Modeling track (\Cref{tab:res-defmod}), while the multitask framework of \citet[BLCU-ICALL]{submissions-blcuicall} yields generally the most consistent performance, it is often outmatched in specific setups. For instance, BLEU-based metrics favor the shared projection technique of \citet[Edinburgh]{submissions-edinburgh} in Russian and French, while the pooling strategies of \citet[IRB-NLP]{submissions-irbnlp} appear especially effective on the Spanish dataset.
As for the Reverse Dictionary track (\Cref{tab:res-revdict}), the strongest contender is generally the Edinburgh team, although the IRB-NLP team almost systematically produces the highest cosine ranking score. Interestingly, BLCU-ICALL, IRB-NLP and Edinburgh all rely on multi-task learning.
Note however that the SGNS targets seem to depict a rather different picture, where the pretraining objectives of \citet[LingJing]{submissions-lingjing} bring about some of the best results.


\section{A deeper dive into our results} \label{sec:discussion}

When looking at the competition results, two trends emerge.
First, the baseline architectures from \Cref{sec:baselines} remain quite competitive with solutions proposed by participants.
Second, scores are generally unsatisfactory, especially in the definition modeling track: we do not see a clear divide between char embeddings and distributional semantic representations. 
The NLG metrics are, in absolute terms, low compared to modern NLP standards and results reported elsewhere on other definition modeling benchmarks.
As for the reverse dictionary track, we see that across all submissions, at least a third of the test set is closer (in terms of cosine distance) to the production than the intended target.

Participants have suggested multiple reasons for these hardships. 
In particular, \citet[MMG]{submissions-mmg} highlight that the automated data compilation in DBnary \citep{serasset-2012-dbnary} is of an unsatisfactory quality. 
Similar remarks can be made with respect to the embeddings, which are trained on rather small corpora.
Other submissions such as \citet[RIGA]{submissions-riga}, \citet[Edinburgh]{submissions-edinburgh}, \citet[IRB-NLP]{submissions-irbnlp} highlight the limited applicability of mainstream NLG metrics, as we ourselves have discussed in \Cref{sec:metrics}.\footnote{See also \citet{mickus-etal-2021-about} for a discussion.}
One last remark is the limited size of our dataset, discussed by the Edinburgh and RIGA teams.
All these remarks suggest avenues for future research: in particular, the release of the full dataset should alleviate some of the concerns with respect to dataset size. 
The MMG team also suggest some concrete preprocessing steps to handle some of the issues they identify in the proposed definitions.

In terms of solutions explored, we can stress that teams have adopted a variety of strategies and architectures: systems used Transformer, RNN and CNN components, often leveraging or exploring multilingualism (\citealt[JSI]{submissions-jsi}; \citealt[Uppsala]{submissions-uppsala}; \citealt[1cademy]{submissions-zhwa3087}; \citealt[BL.research]{submissions-blresearch}), multitasking, or multiple training objectives (\citealt[BLCU-ICALL]{submissions-blcuicall}; 1cadamy; \citealt[IRB-NLP]{submissions-irbnlp}; \citealt[TLDR]{submissions-tldr}; \citealt[Edinburgh]{submissions-edinburgh}). Multi-task training tends to yield varied yet competitive results for our data.
No preponderant architecture emerges from the system descriptions; we note that multiple submissions based their work on other contextualized embedding architectures, trained from scratch on the CODWOE dataset (\citealt[1cademy]{submissions-zhwa3087}; \citealt[LingJing]{submissions-lingjing}).
The comprehensive review of architectures by team 1cadamy suggests nonetheless that Transformers might be less suited to this shared task than recurrent models.

\subsection{Manual analyses}

As for manual evaluations, \citet[BLCU-ICALL]{submissions-blcuicall} provide a thorough review of the errors produced by their model.
\citet[RIGA]{submissions-riga} provide some example outputs of their models, while \citet[TLDR]{submissions-tldr} and \citet[1cademy]{submissions-zhwa3087} include ablation studies.
The most thorough analysis, however, is that of \citet[Edinburgh]{submissions-edinburgh}, who provide both quantitative and qualitative (PCA-based) analyses across embedding architectures, languages, and trial dataset features. 
\citet[IRB-NLP]{submissions-irbnlp} provide an extremely well documented review of their systems performances, along multiple analyses of the embeddings proposed for the shared tasks, ranging from 2D down-projection visualizations to descriptive statistics of components.
We refer the reader to the respective system papers for a more thorough review and focus here on a few promising approaches to summarize trends that emerge from these manual analyses. 

\paragraph{Current metrics are not satisfactory.} The IRB-NLP team highlight that the BLEU scores reported on the shared task are dramatically lower than what is generally expected in the literature; the Edinburgh team even shows that the S-BLEU scores obtained by non-sensical glosses such as ``\texttt{, or .}'' can end up among the highest scores for some languages.
The Reverse Dictionary metrics can also be sensitive to different aspects of the embeddings, as shown by the IRB-NLP team: this can lead to very different rankings of model productions, especially when comparing the cosine-based ranking metric to the cosine and MSE metrics.
BLEU-based scores are also often sensitive to the length of the production, the target, or both, as shown by both the Edinburgh and the Riga teams.

\paragraph{Erroneous productions abound.} Related to the previous remark, many Definition Modeling systems produce irrelevant or under-specified glosses, for which the proposed metrics are not satisfactory. 
For instance, the BLCU-ICALL report 52\% irrelevant glosses and 23.5\% under-specified glosses, from a manual evaluation of 200 productions. 
Other participating teams, such as RIGA or IRB-NLP, also display generated glosses with varying degress of semantic accuracy.

\paragraph{Embeddings contain more than semantics.} The Edinburgh team highlights how different linguistic features retrieved from the trial dataset can significantly impact the scores they observe. They also highlight that char embeddings are separable by length, and that the Electra embeddings are clustered according to their frequency. 

\paragraph{Not all setups are created equal.} The Uppsala team report that Russian seems to be the most effective data source in their multilingual transfer experiments.
The IRB-NLP team stresses that vector component distributions across languages and architectures as well as gloss length across languages can take very different values, and they also include 2D visualization suggesting the Electra embeddings tend to form neat cluster not observed for SGNS embeddings. 
Scores also vary quite a lot across setups (cf. \Cref{tab:res-defmod,tab:res-revdict}).

\section{Conclusions and future perspectives}
The CODWOE shared task was constructed so that participants' submissions would be likely to have linguistic significance.
Yet, it is not trivial to tease apart the various factors that lead to the overall low results we observed.
While the inadequacy of mainstream NLG metrics and the limitations of the dataset certainly play a role, they do not resolve the fundamental issue that we wished to investigate with CODWOE.
Whether word embeddings and dictionaries contain the same information is not a solved research problem.

This has two immediate consequences: firstly, one can question the use of definition modeling as an evaluation tool for embeddings, as suggested by the seminal work of \citet{DBLP:conf/aaai/NorasetLBD17}.
The CODWOE shared task results indicate that the metrics currently used in the field are rife with caveats; in the controlled setup we have proposed here, participants rarely, if ever, found that character-based embeddings starkly contrasted with distributional semantic representations. 

Second, one can question whether definition modeling and reverse dictionary are fit for building lexical resources for under-tooled languages: the crosslingual route proposed by \citet{bear-cook-2021-cross} seems more practical than training models from scratch, even with relatively large datasets.
Our embeddings were trained on corpora comparable in size to the 1B Words benchmark \citep{DBLP:journals/corr/ChelbaMSGBK13}: while modern text corpora are now several orders of magnitude larger, this dataset remained a landmark for several years.
Our definitions were selected from DBnary \citep{serasset-2012-dbnary}, which focuses the largest Wiktionary projects.

Overall, the CODWOE shared task has been a success: we were able to show that the task at hand was far from trivial and we drew significant interest towards the issues addressed in the Definition Modeling and Reverse Dictionary literature.
In future work, we plan to investigate better ways to perform NLG evaluation
for the Definition Modeling task (in particular relying on human annotations) and we plan to focus on existing embeddings trained from very large corpora.

\section*{Acknowledgements}
We thank the annotators who have helped us in putting an annotated dataset in a very short amount of time: Elena del Olmo Suárez, Hermès Martinez, Maria Copot, Nikolay Chepurnykh and Toma Gotkova, as well as Cyril Pestel for his help with data hosting.

This work was also supported  by a public grant overseen by the French National Research Agency (ANR) as part of the ``Investissements d'Avenir'' program: Idex \emph{Lorraine Universit\'e d'Excellence} (reference: ANR-15-IDEX-0004).

\bibliography{anthology,custom,submissions}
\bibliographystyle{acl_natbib}

\appendix

\end{document}